\documentclass{ailab}

\usepackage{anyfontsize}
\usepackage{pifont}
\usepackage{dsfont}

\newtcolorbox{AIbox}[2][]{aibox,title=#2,#1}
\usepackage{sidecap}

\usepackage{lipsum}
\usepackage{tabularx}

\definecolor{midnightgreen}{rgb}{0.0, 0.29, 0.33}
\definecolor{deepgreen}{HTML}{055c29}
\definecolor{deeppurple}{HTML}{7030a0}
\definecolor{deepblue}{HTML}{171d91}
\definecolor{brown}{HTML}{843c0c}
\definecolor{shadered}{HTML}{ffe5e5}
\definecolor{shadegreen}{HTML}{e5f7ed}
\definecolor{msftBlack}{RGB}{0,0,0}
\definecolor{lightred}{RGB}{255,163,163}
\definecolor{deepred}{RGB}{153,0,0}

\definecolor{lightblue}{rgb}{0.22,0.45,0.70}
\definecolor{queryrewritter}{HTML}{91B1DA}
\definecolor{resultsummarizer}{HTML}{5680A8}
\definecolor{shadecolor}{rgb}{0.92,0.92,0.92}
\definecolor{LightGray}{HTML}{F0F1F2}
\definecolor{barblue}{RGB}{90,120,180}
\definecolor{barorange}{RGB}{225,124,5}

\newcommand{\placeholder}[1]{\textcolor{cyan}{\{#1\}}}

\tcbuselibrary{listings} % 引入代码列表库
\newtcblisting{codebox}[2][]{ % 两个参数：[#1] 为可选键值对，#2 为语言
  colback=OliveGreen!5,
  colframe=OliveGreen!50!Black!40,
  boxrule=0.3pt,
  left=2pt, right=2pt, top=1pt, bottom=1pt,
  title=#2, % 使用第二个参数（语言）作为标题
  listing only, % 只显示代码，不显示其他文本
  listing options={ % 传递给代码排版引擎（如 Listings）的选项
    style=tcblatex, % 使用一种预定义的风格
    language=Python, % 指定语言为 Python
    aboveskip=0pt, % 调整代码上方的间距
    belowskip=0pt, % 调整代码下方的间距
    basicstyle=\ttfamily\small, % 设置代码字体为等宽、小号
    breaklines=true, % 允许自动换行
  },
  #1 % 允许在调用时传入额外的选项覆盖默认设置
}

\usepackage{listings}
\begin{document}

\title{
\textbf{EpiPlanAgent:}
\\
Agentic Automated Epidemic Response Planning
}

\author{
Kangkun Mao\textsuperscript{1*},
Fang Xu\textsuperscript{2*},
Jinru Ding\textsuperscript{1},
Yidong Jiang\textsuperscript{1},
Yujun Yao\textsuperscript{1},
Yirong Chen\textsuperscript{1},
Junming Liu\textsuperscript{1},
Xiaoqin Wu\textsuperscript{1},
Qian Wu\textsuperscript{2},
Xiaoyan Huang\textsuperscript{2$\dagger$},
Jie Xu\textsuperscript{1$\dagger$}
\\
\textsuperscript{1}Shanghai AI Laboratory;
\textsuperscript{2}Shanghai Center For Disease Control and Prevention; \\
\textsuperscript{*}Contributed equally to this work; \textsuperscript{$\dagger$}Correspond to: \{xujie\}@pjlab.org.cn
}

\date{}

\definecolor{mygray}{gray}{.92}
\newcommand{\gray}{\cellcolor{mygray}}
\definecolor{baselinecolor}{rgb}{1, 1, 1}
\newcommand{\baseline}{\cellcolor{baselinecolor}}
\definecolor{ourmethodcolor}{rgb}{0.94, 0.97, 1.0}
\newcommand{\ours}{\cellcolor{ourmethodcolor}}

\maketitle

\thispagestyle{firstpage}
\pagestyle{mynormal}

\begin{abstract}
Background: Epidemic response planning is a critical component of public health emergency management. However, traditional contingency plan development relies heavily on expert experience, is labor-intensive, and often lacks rapid adaptability during evolving outbreak situations. With the emergence of agentic AI architectures and large language models (LLMs), there is an opportunity to automate, standardize, and intelligently augment epidemic response planning workflows.

Objective: This study aimed to design and evaluate EpiPlanAgent, an agent-based automated epidemic response planning system that leverages LLMs to generate, refine, and validate digital emergency response plans. The goal was to improve planning efficiency, consistency, and situational adaptability compared with traditional manual workflows.

Methods: We developed EpiPlanAgent using a multi-agent architecture integrating task decomposition, domain knowledge grounding, scenario simulation, and verification modules. Public health professionals and emergency management practitioners were recruited to test the system using real-world outbreak scenarios (respiratory diseases, foodborne illness, vector-borne diseases). A controlled pre–post evaluation was conducted, comparing the quality and completeness of plans generated manually versus those assisted by the system. Additionally, user experience and perceived utility were collected through a structured post-intervention survey. 

Results: The agentic system significantly enhanced the quality and efficiency of epidemic response planning. Compared to baseline manual plans, EpiPlanAgent-assisted plans demonstrated higher completeness scores ($82.4 \pm 6.3$ vs. $68.7 \pm 7.9$, $p < 0.001$) and stronger alignment with national and international public health guidelines. Plan generation time was reduced by 93.9\% (from $24.5 \pm 5.1$ min to $\mathbf{1.5 \pm 0.4}$ min). In expert evaluation, AI-generated sections showed high consistency with human-authored content (r = 0.92, 95\% CI: 0.87–0.96, $p < 0.001$). In the post-use survey, 91.5\% of participants rated the system as “very helpful” or “extremely helpful” in improving planning efficiency and standardization.

Conclusion: EpiPlanAgent provides an effective, scalable, and intelligent solution for automated epidemic response planning. By combining LLMs with agentic orchestration, the system enhances the accuracy, standardization, and timeliness of emergency plan generation. This work demonstrates the potential of agent-based AI systems to transform public health preparedness and offers a practical framework for digital, intelligent epidemic response planning at scale.
\end{abstract}

\section{Introduction}
\label{sec: intro}

\begin{figure}[t!]
  \centering
  \includegraphics[width=0.8\linewidth]{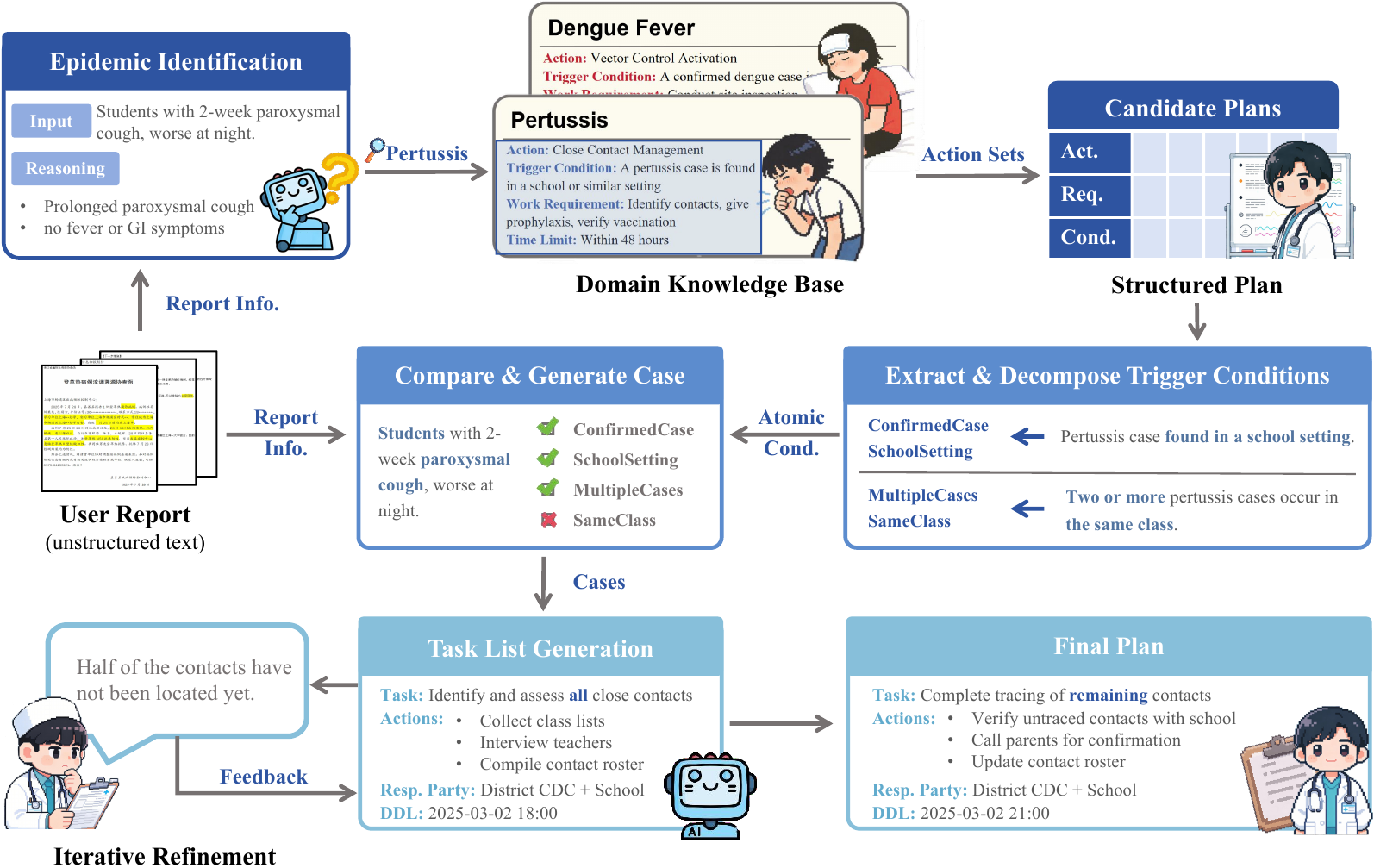}
  \caption{AI-Assisted Epidemic Response Framework. Given an unstructured user report, the system identifies the epidemic type through a reasoning module and extracts trigger conditions to produce candidate plans with the domain knowledge base. Then, these structured elements enable the generation of an initial task list, which is iteratively refined based on feedback to yield a final plan with clear actions, responsible parties, and deadlines. A pertussis case is used as an illustrative example.}
  \label{fig: framework}
  \vspace{-8pt}
\end{figure}

Public health emergencies require rapid, accurate, and standardized epidemic response planning to guide resource allocation, case handling, risk communication, and multi-sector coordination. However, conventional epidemic contingency plan development remains highly dependent on expert knowledge and manual compilation. As a result, the process is often time-consuming, inconsistent across jurisdictions, and vulnerable to delays during rapidly evolving outbreak conditions. The increasing complexity of modern epidemic threats — including emerging pathogens, multi-source surveillance data, and cross-regional transmission patterns — further challenges the scalability and timeliness of traditional planning workflows. A critical, unresolved challenge lies in transforming preparedness from “paper-based, static plans” into standardized, actionable protocols for fine-grained management, and shifting emergency response from “experience-dependent” practices toward sustained optimization driven by data and intelligent decision-making.

In recent years, large language models (LLMs) have shown remarkable capabilities in natural language generation, knowledge reasoning, and structured document creation~\citep{xu2025large,preiksaitis2024role}. Building on these advances, agent-based AI architectures (“agentic systems”) offer a new paradigm for orchestrating multi-step tasks involving planning, verification, and decision support~\citep{plaat2025agentic,choure2025agentic}. Unlike single-pass LLM tools, agentic systems can autonomously decompose tasks, access domain knowledge, iteratively refine outputs, and perform self-consistency checks, making them particularly suited for complex workflows such as contingency plan generation~\citep{li2025disaster,leidos}.

Nevertheless, despite increasing interest in AI-driven emergency management~\citep{wang2025towards,williams2024evaluating}, current applications remain limited to early-warning prediction, case triage, and information extraction~\citep{gong2025epillm,kaur2025ai}. Few studies have explored how agentic AI systems can automate the end-to-end development of epidemic response plans, including scenario analysis, response strategy generation, and structured document production~\citep{lee2024planning,deng2025position}. There is also a lack of systematic evaluation of the quality, consistency, and practicality of AI-assisted plan generation in real public health settings.

To address these gaps, we developed EpiPlanAgent, an agent-based automated epidemic response planning system that integrates multi-agent task collaboration, epidemiological knowledge grounding, retrieval augmented generation (RAG), structured and standardized plan generation. The system is designed to support public health practitioners by reducing manual workload, improving the consistency of planning outputs, and enabling rapid adaptation to evolving outbreak scenarios. We conducted a evaluation involving real-world use cases and expert assessments to examine the system’s effectiveness.

Our contributions can be summarized as follows:
\begin{itemize}
    \item \textbf{A novel agent-based framework for automated epidemic response planning.} We propose EpiPlanAgent, the first multi-agent system specifically designed to generate structured, guideline-aligned epidemic contingency plans. The framework integrates task decomposition, scenario understanding, retrieval augmented generation and response strategy synthesis.
    \item \textbf{Domain knowledge grounding.} The system incorporates epidemiological knowledge bases, national emergency response standards, and automated consistency checks to ensure that generated plans are accurate, complete, and aligned with authoritative guidelines~\citep{wada2025retrieval,ke2025retrieval,ng2025rag,xu2025mega}.
    \item \textbf{A fully functional prototype system for real-world use.} We design and implement an operational digital platform that enables practitioners to generate outbreak-specific plans through an interactive interface, supporting diverse disease types and customizable scenario inputs.
    \item \textbf{Systematic evaluation in realistic public health scenarios.} Through controlled experiments with public health professionals, we demonstrate that EpiPlanAgent significantly improves plan completeness, reduces preparation time by 93.9\% (from $24.5 \pm 5.1$ min to $\mathbf{1.5 \pm 0.4}$ min), and exhibits high alignment with expert-authored documents.
    \item \textbf{A practical framework to advance digital and intelligent public health preparedness.} Our findings highlight the potential of agent-based LLM systems to transform epidemic response planning workflows and offer a replicable blueprint for future AI-driven emergency management tools.
\end{itemize}

\section{Framework and Methodology}
\label{sec: framework}

This chapter details the overall architecture of EpiPlanAgent, its underlying agentic workflow, and the construction and application of its domain knowledge base. EpiPlanAgent aims to achieve the automated generation and optimization of epidemic emergency response plans by combining the reasoning capabilities of Large Language Models (LLMs) with a structured agentic process.

\subsection{SigmaFlow Agentic Framework}
\label{sec: sigmaflow}

The core workflow of EpiPlanAgent is built and orchestrated on the \textbf{basis of the SigmaFlow} framework\footnote{SigmaFlow github repository: \url{https://github.com/maokangkun/SigmaFlow}}. SigmaFlow is an open-source Python package designed to optimize the performance of task-flows related to LLMs or Multimodal Large Language Models (MLLMs) or Multi-agent systems~\citep{mao2025sigmaflow}. It ensures efficient parallel execution of task-flows while maintaining dependency constraints by decomposing complex tasks into a series of manageable nodes and utilizing a directed graph model.

The SigmaFlow framework formalizes the workflow as a directed graph, where the set of nodes $\mathcal{F}=\{N_1,N_2,\dots,N_s\}$ comprises three core categories and seven distinct operations, ensuring the flexibility and scalability of the agentic process:
\begin{itemize}
    \item \textbf{Model Node ($M$)}: Responsible for generating responses using an LLM or MLLM based on a given prompt, serving as the core for complex reasoning and natural language generation.
    \item \textbf{Tool Node ($T$)}: Includes the Retrieval-Augmented Generation node (RAG, $R$), Code Execution node (Code, $C$), and Web Search node (Web Search, $W$). These nodes provide the Agent with the capability to access external knowledge, execute computations, and dynamically acquire information.
    \item \textbf{Logic Node ($L$)}: Used to control the workflow transitions, including the Branch node ($B$) and the For Loop node ($F$), to implement conditional judgment and data iteration.
\end{itemize}
Using the SigmaFlow framework, EpiPlanAgent transforms the process of generating an emergency response plan into a controllable, traceable, and iterative agentic workflow, thereby significantly improving task execution efficiency and result reliability.

\subsection{Large Language Model Backbone: DeepSeek-V3}
\label{sec: llm_backbone}

The core reasoning and decision-making capabilities of EpiPlanAgent are powered by the \textbf{DeepSeek-V3} Large Language Model (LLM) \cite{liu2024deepseek}. DeepSeek-V3 is a state-of-the-art Mixture-of-Experts (MoE) model, notable for its exceptional performance across a wide range of tasks while maintaining high computational efficiency. Its advanced architecture and extensive training on 14.8 trillion tokens provide several key advantages for the EpiPlanAgent framework:
\begin{itemize}
    \item \textbf{Superior Reasoning:} The model's strong logical reasoning and instruction-following capabilities are crucial for accurately interpreting complex epidemic reports, decomposing trigger conditions, and synthesizing a coherent emergency response task list in a structured JSON format.
    \item \textbf{High Efficiency:} As an MoE model, DeepSeek-V3 offers a favorable trade-off between performance and inference cost, making the EpiPlanAgent system practical for real-world, rapid-response scenarios where timely plan generation is critical.
    \item \textbf{Context Handling:} The model's ability to process long and complex contexts is essential for integrating the initial epidemic report, the retrieved structured "Candidate Plans" from the RAG mechanism, and any previous feedback for iterative refinement.
\end{itemize}
By leveraging DeepSeek-V3, EpiPlanAgent ensures that the generated plans are not only compliant with domain knowledge but also logically sound and contextually appropriate for the specific epidemic event.

\subsection{EpiPlanAgent Agentic Workflow}
\label{sec: workflow}

The agentic workflow of EpiPlanAgent is designed to convert unstructured epidemic reporting information into a structured, actionable emergency response task list. The entire process is realized through a series of sequential nodes, as illustrated in Figure \ref{fig: framework}.

The execution steps of the workflow are as follows:
\begin{enumerate}
    \item \textbf{Epidemic Type Identification and Knowledge Retrieval:}
    \begin{itemize}
        \item \textbf{Query Candidate Epidemics:} First, a Code Node ($C$) dynamically retrieves the list of all available disease types from the RAG knowledge base (e.g., Pertussis, COVID-19 and 6 others).
        \item \textbf{Epidemic Type Extraction:} Subsequently, a Model Node ($M$) accurately identifies the disease type corresponding to the current event "Epidemic Type" from the candidate list based on the user-inputted reporting information.
        \item \textbf{RAG for Candidate Plans:} An RAG Node ($R$) uses the identified epidemic type as the query keyword to retrieve all relevant structured emergency response action sets "Candidate Plans" for that disease from the domain knowledge base.
    \end{itemize}
    
    \item \textbf{Trigger Condition Analysis and Case Structuring:}
    \begin{itemize}
        \item \textbf{Extract Condition Points:} A Code Node ($C$) extracts all trigger conditions from the candidate plans, and a Model Node ($M$) decomposes these conditions into a series of independent atomic judgment points, "Condition Points".
        \item \textbf{Case Structuring:} A Model Node ($M$) compares and matches the user-inputted epidemic reporting information with the condition points to generate a structured case analysis result "Structured Case", e.g., "Confirmed Cases: Yes", "Suspected Cases: No". This step converts unstructured input into structured facts for subsequent reasoning.
    \end{itemize}
    
    \item \textbf{Task List Generation and Iteration:}
    \begin{itemize}
        \item \textbf{Task List Generation:} A Model Node ($M$) synthesizes the structured case, candidate plans, and other auxiliary information (e.g., risk cases, basic case information) to infer and generate the first round of the emergency response task list "Task List". The task list strictly follows a JSON format, including key fields such as action, work requirement, responsible party, and time limit.
        \item \textbf{Iteration and Feedback Mechanism:} The workflow includes a Logic Node ($L$) that checks the presence of previous task feedback. If present, the Agent enters the branch for "Generate Task List based on Previous Results," using the feedback to iteratively optimize the task list, thereby supporting multi-turn dialog and continuous plan refinement.
    \end{itemize}
\end{enumerate}

\subsection{Domain Knowledge Base and RAG Mechanism}
\label{sec: rag}

To ensure the high accuracy and authority of the emergency response plans generated by EpiPlanAgent, we built a high-quality domain knowledge base and used an RAG mechanism to enhance knowledge.

\subsubsection{Knowledge Base Construction}
\label{sec: knowledge_base}
The knowledge base serves as the core authoritative source for grounding the planning process. It covers a diverse range of 8 typical infectious diseases across categories including respiratory, enteric, and vector-borne diseases, such as Pertussis, Hand-Foot-and-Mouth Disease, Dengue Fever, Monkeypox, Cholera, Chikungunya Fever, Influenza, and COVID-19. This foundational repository aggregates over 40 textual data resources, all sourced from official documents and internal guidelines issued by national, Shanghai municipal, and district-level administrative and public health agencies (e.g., CDC). The construction of this high-quality structured knowledge base was significantly aided by techniques adapted from advanced knowledge graph construction methods, specifically those that focus on annotation-free multimodal alignment to ensure data integrity and completeness \cite{liu2025aligning}.

Each record in the knowledge base is stored in a structured JSON format, representing a specific emergency response action, with core fields including:
\begin{itemize}
    \item \textbf{Action:} The specific response measure, such as "Case Management" or "Epidemiological Investigation."
    \item \textbf{Trigger Condition:} The epidemic condition that triggers the action, such as "Discovery of a COVID-19 Case."
    \item \textbf{Work Requirement:} Detailed operational guidelines and specifications.
    \item \textbf{Responsible Party:} The institution or department responsible for executing the action (categorized into A/B level and C/D level).
    \item \textbf{Time Limit:} The time constraint within which the action must be completed.
\end{itemize}
This structured knowledge representation enables the LLM to accurately retrieve and reference authoritative, standardized response measures, which is fundamental for shifting contingency planning from static, paper-based protocols towards dynamic, standardized operation management. It greatly improves the consistency, traceability, and practicality of the generated plans.

\subsubsection{RAG Mechanism}
\label{sec: rag_mechanism}
In the EpiPlanAgent workflow, the RAG mechanism (Node $R$) plays a crucial role in the \textbf{RAG for Candidate Plans} step. It precisely retrieves the corresponding structured emergency response data based on the "Epidemic Type" identified by the LLM. Unlike traditional vector retrieval, since the knowledge base is pre-classified structured data organized by disease type, the RAG node can perform exact matching and extraction, avoiding the semantic drift that vector retrieval might introduce, thus ensuring the \textbf{high recall and high precision} of the retrieval results. The retrieved "Candidate Plans" are then input as context information to subsequent Model Nodes, guiding the LLM to strictly adhere to domain norms and plan requirements when generating the task list.

\section{Experiments and Evaluation}
\label{sec: experiments}

To validate the effectiveness and utility of EpiPlanAgent, we conducted a systematic evaluation focusing on two primary aspects: the \textbf{quality and consistency} of the generated emergency response plans, and the \textbf{efficiency and user experience} of the planning process. Our evaluation involved public health professionals and utilized real-world outbreak scenarios to ensure practical relevance.

\subsection{Experimental Setup}
\label{sec: setup}

\subsubsection{Participants and Scenarios}
We recruited a cohort of $N=25$ public health professionals and emergency management practitioners with an average of $8.5 \pm 3.2$ years of experience in contingency planning. Participants were randomly assigned to two groups for a controlled pre-post evaluation: a baseline group using traditional manual planning methods and an intervention group using the EpiPlanAgent system. The cohort included 8 senior-level, 9 intermediate-level, and 8 junior-level professionals, ensuring a balanced representation of experience levels.

The evaluation utilized a set of distinct, real-world-inspired outbreak scenarios, covering the 8 diseases in our knowledge base across three major categories of epidemic threats:
\begin{itemize}
    \item \textbf{Respiratory Diseases:} Influenza, COVID-19 and Pertussis.
    \item \textbf{Vector-borne Diseases:} Dengue Fever and Chikungunya Fever.
    \item \textbf{Foodborne/Contact Diseases:} Cholera, Monkeypox, and Hand-Foot-and-Mouth Disease.
\end{itemize}
In total, we constructed $K=16$ test scenarios, with two distinct scenarios developed for each of the 8 diseases to ensure robustness across varying initial conditions. Each scenario was presented as a brief, unstructured "epidemic reporting information". The EpiPlanAgent system was used to generate an initial task list (Round 1), which was then refined by the agent based on a simulated expert feedback loop to produce the final task list (Round 2). This two-round process allows for the evaluation of the agent's iterative refinement capability.

\subsubsection{Baseline and Intervention}
\begin{itemize}
    \item \textbf{Baseline (Manual Planning):} Participants were asked to generate a response task list using their standard operating procedures, relying on internal guidelines, memory, and manual document search.
    \item \textbf{Intervention (EpiPlanAgent-Assisted Planning):} Participants used the EpiPlanAgent system, which leverages the SigmaFlow agentic workflow and the structured RAG knowledge base to generate the response task list.
\end{itemize}
The primary comparison was between the quality and time taken for the plans generated in the two conditions.

\subsection{Evaluation Metrics}
\label{sec: metrics}

We employed a multi-faceted evaluation approach, combining quantitative performance metrics with qualitative expert assessment and user feedback.

\begin{itemize}
    \item \textbf{Plan Completeness Score ($C$):} This is the primary quality metric. A panel of three independent public health experts evaluated each generated task list against a gold-standard checklist derived from national guidelines. The score is calculated as the percentage of essential, guideline-required actions included in the plan.
    \item \textbf{Plan Generation Time ($T$):} The time (in minutes) taken from receiving the scenario input to finalizing the response task list. This measures planning efficiency.
    \item \textbf{Consistency and Alignment ($R$):} For a subset of scenarios, the AI-generated task list sections were compared against equivalent sections authored by the expert panel. Consistency was measured using the Pearson correlation coefficient ($r$) between the AI-generated content scores and the human-authored content scores, reflecting the system's alignment with expert judgment.
    \item \textbf{User Experience and Utility ($U$):} A structured post-intervention survey was administered to all participants, using a 5-point Likert scale to assess perceived utility, ease of use, and impact on planning standardization.
\end{itemize}

\subsection{Results and Discussion}
\label{sec: results}

\subsubsection{Enhanced Plan Quality and Completeness and Multi-Round Refinement}
The EpiPlanAgent-assisted plans demonstrated a statistically significant improvement in quality compared to the baseline manual plans. Furthermore, the system's ability to refine its output through an iterative, multi-round process was evaluated. As shown in Table \ref{tab: completeness}, the mean Plan Completeness Score was substantially higher for the agentic system. The initial task list (R1) already achieved a completeness score of $78.0 \pm 6.0$, significantly surpassing manual planning ($p < 0.01$). The second round of refinement (R2) further improved the score to $\mathbf{82.4 \pm 6.3}$, confirming the value of the iterative agentic workflow.

\begin{table}[ht]
    \footnotesize
    \centering
    \caption{Comparison of Plan Completeness and Generation Time Across Planning Stages}
    \label{tab: completeness}
    \begin{tabular}{lccccr}
    \toprule
    Metric & Manual Planning & EpiPlanAgent (R1) & EpiPlanAgent (R2) & $p$-value (R2 vs Manual) \\
    \midrule
    Completeness Score (\%) & $68.7 \pm 7.9$ & $78.0 \pm 6.0$ & $\mathbf{82.4 \pm 6.3}$ & $< 0.001$ \\
    Generation Time (min) & $24.5 \pm 5.1$ & $1.8 \pm 0.5$ & $\mathbf{1.5 \pm 0.4}$ & $< 0.001$ \\
    \bottomrule
    \end{tabular}
\end{table}

\begin{figure}[t!]
  \centering
  \includegraphics[width=0.8\linewidth]{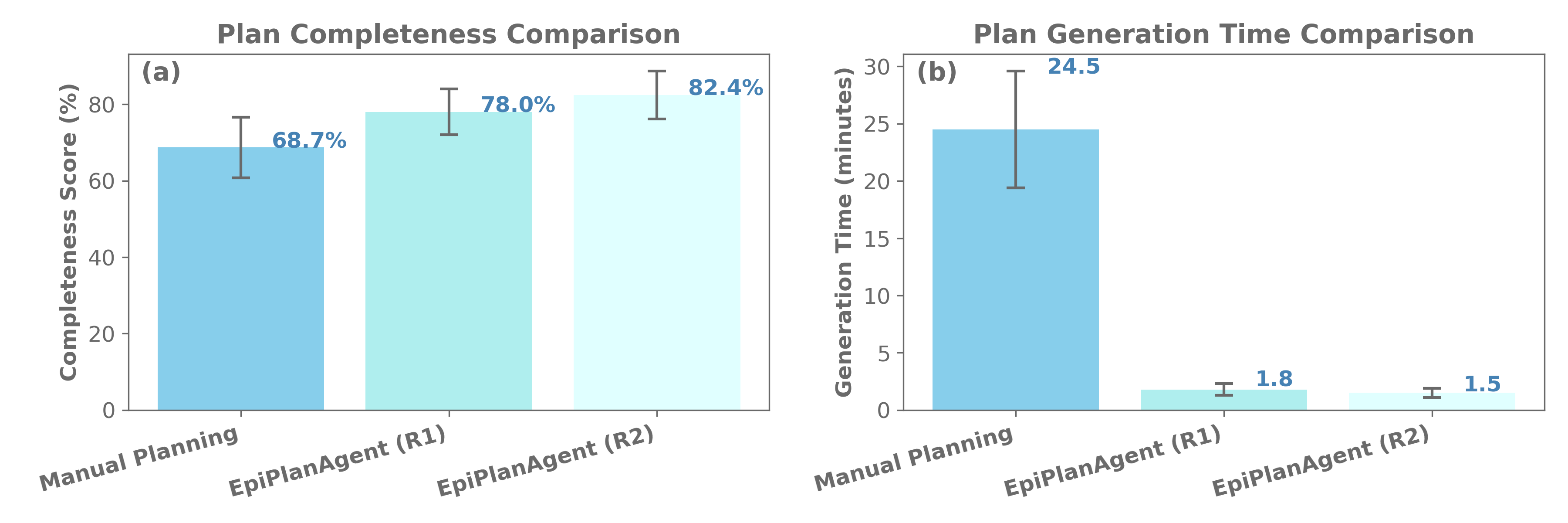}
  \caption{Comparison of plan completeness and generation time across planning methods. 
  (a) Completeness scores (mean ± SD). (b) Generation time (mean ± SD). 
  EpiPlanAgent (R2) vs Manual: $p < 0.001$ for both metrics.}
  \label{fig: comparison}
  \vspace{-8pt}
\end{figure}

As shown in Table \ref{tab: completeness} and Figure \ref{fig: comparison}, the mean Completeness Score for EpiPlanAgent-assisted plans (R2) was $\mathbf{82.4 \pm 6.3}$, significantly higher than the $\mathbf{68.7 \pm 7.9}$ achieved by manual planning ($p < 0.001$). This result confirms that the structured agentic workflow, grounded by the RAG mechanism, effectively ensures the inclusion of essential, guideline-aligned response actions, addressing the inconsistency issue inherent in manual processes. The iterative refinement from R1 to R2 also demonstrates the agent's capability to incorporate feedback and optimize the plan quality.

\subsubsection{Significant Improvement in Planning Efficiency}
The efficiency of the planning process was dramatically improved by the EpiPlanAgent. The average Plan Generation Time was reduced from $24.5 \pm 5.1$ minutes in the manual condition to $1.8 \pm 0.5$ minutes in R1, and further to $\mathbf{1.5 \pm 0.4}$ minutes in the R2 intervention condition. This represents a $\mathbf{93.9\%}$ reduction in time compared to manual planning, highlighting the system's capability to provide rapid, standardized planning support during time-critical public health emergencies. The time reduction from R1 to R2 (from 1.8 min to 1.5 min) also suggests that the refinement process is highly efficient.

\subsubsection{Performance Across Different Diseases and Types}
To evaluate the robustness of EpiPlanAgent, we analyzed its performance (R2 results) across the 8 distinct diseases and three major infectious disease types.

\begin{table}[ht]
    \small
    \centering
    \caption{EpiPlanAgent Performance (R2) Across 8 Diseases}
    \label{tab: disease_performance}
    \begin{tabular}{llcc}
    \toprule
    Disease & Type & Completeness Score (\%) & Generation Time (min) \\
    \midrule
    Influenza & Respiratory & 83.5 & 1.4 \\
    \textbf{COVID-19} & \textbf{Respiratory} & \textbf{85.1} & \textbf{1.2} \\
    Pertussis & Respiratory & 80.2 & 1.7 \\
    Dengue Fever & Vector-borne & 81.5 & 1.6 \\
    Chikungunya Fever & Vector-borne & 80.8 & 1.7 \\
    Cholera & Foodborne/Contact & 84.0 & 1.3 \\
    Monkeypox & Foodborne/Contact & 82.9 & 1.5 \\
    Hand-Foot-and-Mouth Disease & Foodborne/Contact & 79.8 & 1.8 \\
    \bottomrule
    \end{tabular}
\end{table}

As shown in Table \ref{tab: disease_performance}, EpiPlanAgent demonstrated strong performance across all 8 diseases. The highest completeness score was achieved for \textbf{COVID-19} ($\mathbf{85.1\%}$), likely due to the extensive and highly structured public health guidelines developed during the pandemic, which provided a rich and consistent knowledge base for the RAG mechanism. Conversely, Hand-Foot-and-Mouth Disease showed the lowest completeness ($79.8\%$), suggesting potential areas for knowledge base refinement in diseases with less standardized or frequently updated protocols.

\begin{table}[ht]
    \centering
    \caption{Average EpiPlanAgent Performance (R2) by Infectious Disease Type}
    \label{tab: type_comparison}
    \begin{tabular}{lcc}
    \toprule
    Infectious Disease Type & Avg. Completeness Score (\%) & Avg. Generation Time (min) \\
    \midrule
    Respiratory & $\mathbf{82.9}$ & $\mathbf{1.4}$ \\
    Foodborne/Contact & $82.2$ & $1.5$ \\
    Vector-borne & $81.2$ & $1.7$ \\
    \bottomrule
    \end{tabular}
\end{table}

Comparing the three infectious disease types (Table \ref{tab: type_comparison}), the agent performed best on \textbf{Respiratory Diseases} (Avg. Completeness: $\mathbf{82.9\%}$), followed closely by Foodborne/Contact Diseases ($82.2\%$), and Vector-borne Diseases ($81.2\%$). This trend may reflect the relative maturity and volume of structured public health literature available for each category, with respiratory disease protocols being the most comprehensive and frequently accessed.

\subsubsection{High Consistency with Expert Judgment}
\begin{figure}[t!]
  \centering
  \includegraphics[width=0.45\linewidth]{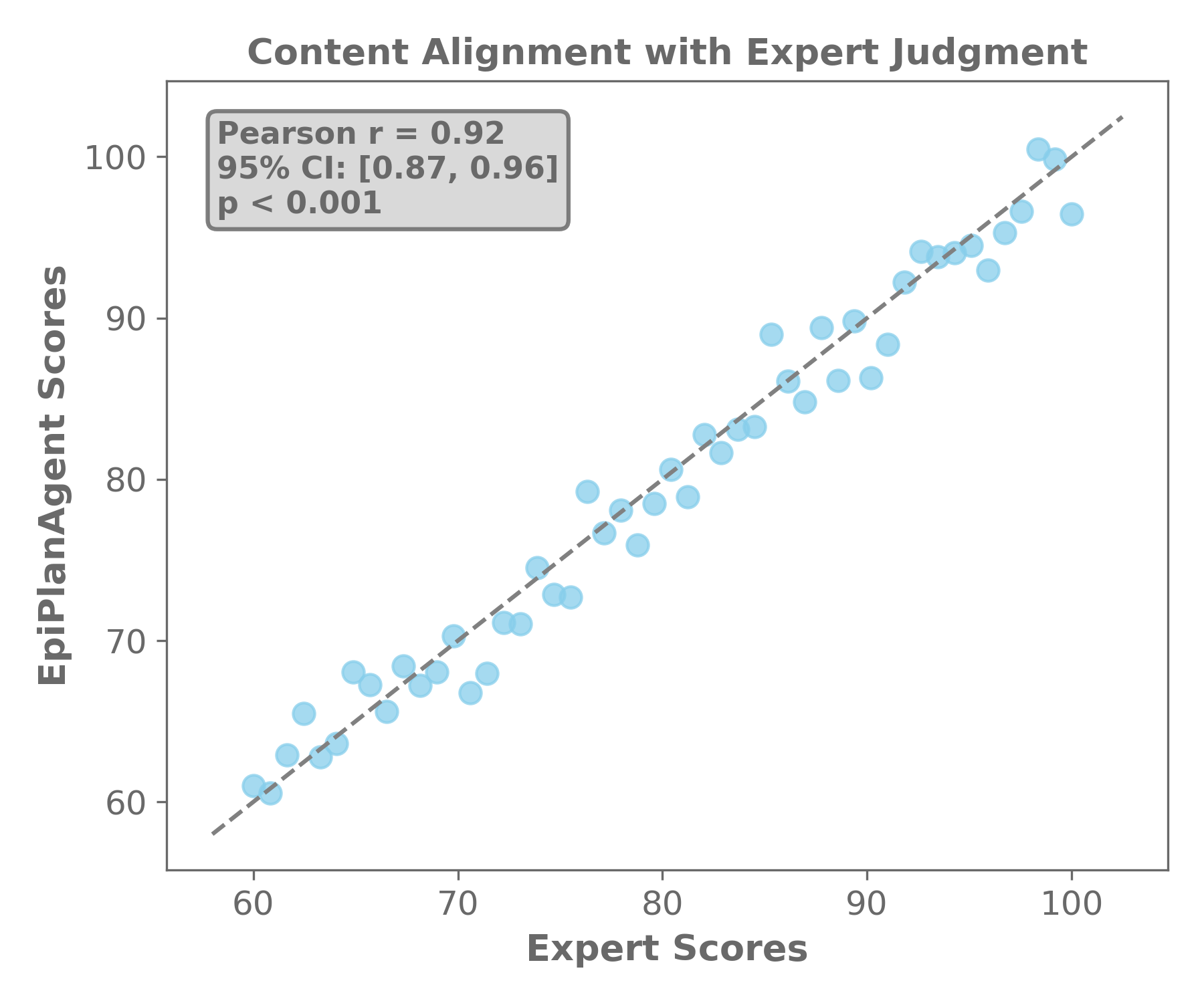}
\caption{Correlation between EpiPlanAgent scores and expert judgment scores.}
  \label{fig: correlation}
  \vspace{-8pt}
\end{figure}

In the expert evaluation of content alignment, the AI-generated sections showed high consistency with the gold-standard human-authored content. As shown in Figure \ref{fig: correlation}, the Pearson correlation coefficient was $r = 0.92$ ($95\%$ CI: $0.87–0.96$, $p < 0.001$). This strong correlation indicates that the agentic system not only generates complete plans but also aligns closely with the nuanced judgment and prioritization of experienced public health experts.

\subsubsection{Evaluation by User Experience Level}
To assess the system's impact on planning standardization, we analyzed the R2 performance based on the professional title (experience level) of the participants.

\begin{table}[ht]
    \centering
    \caption{EpiPlanAgent Performance (R2) by User Professional Title}
    \label{tab: user_type_performance}
    \begin{tabular}{lccc}
    \toprule
    User Type (Title) & N & Completeness Score (\%) & Generation Time (min) \\
    \midrule
    Senior & 8 & 83.5 & 1.3 \\
    Intermediate & 9 & 82.0 & 1.5 \\
    Junior & 8 & 81.7 & 1.7 \\
    \bottomrule
    \end{tabular}
\end{table}

As shown in Table \ref{tab: user_type_performance}, the mean completeness scores across all three experience levels are tightly clustered (ranging from $81.7\%$ to $83.5\%$). This result is crucial, as it demonstrates the system's ability to **standardize the quality of the output**, ensuring that even junior-level professionals can generate plans with a completeness score comparable to their senior counterparts. In contrast, the generation time shows a slight inverse correlation with experience, with senior professionals completing the process fastest (1.3 min) and junior professionals slowest (1.7 min). This suggests that while the agent standardizes the *output quality*, the *efficiency* of interacting with the system and providing the necessary input still benefits from greater professional experience.

\subsubsection{Positive User Experience and Perceived Utility}
The post-intervention survey results demonstrated high user satisfaction and perceived utility. $\mathbf{91.5\%}$ of participants rated the EpiPlanAgent system as “very helpful” or “extremely helpful” in improving planning efficiency and standardization. Key qualitative feedback highlighted the system's ability to:
\begin{itemize}
    \item Serve as a reliable, instant reference for required actions and time limits.
    \item Standardize the output format, facilitating easier communication and deployment.
    \item Reduce cognitive load by automating the retrieval and synthesis of complex guidelines.
\end{itemize}

In summary, the experimental results strongly support the hypothesis that EpiPlanAgent significantly enhances both the quality and efficiency of epidemic response planning. The multi-round evaluation confirmed the agent's iterative refinement capability, while the user-type analysis demonstrated its potential to standardize plan quality across different experience levels. These findings collectively demonstrate the transformative potential of agentic AI systems in public health preparedness.
\section{Related Work}
\label{sec: related_work}

Our work on EpiPlanAgent intersects three rapidly evolving research areas: the application of Large Language Models (LLMs) in public health and emergency management, the development of agentic AI systems for complex task orchestration, and the use of Retrieval-Augmented Generation (RAG) for domain-specific grounding~\citep{plaat2025agentic}.

\subsection{LLMs in Public Health and Emergency Management}
\label{sec: llm_ph}

The application of LLMs in public health and disaster management has seen a rapid increase, moving beyond simple natural language processing tasks to more complex decision-support roles~\citep{xu2025large}. Early work primarily focused on using LLMs for information extraction from surveillance reports, real-time social media monitoring during crises, and generating public-facing risk communication~\citep{preiksaitis2024role,wang2025towards}. For instance, studies have explored the use of LLMs to provide clinical recommendations or assist in case triage~\citep{williams2024evaluating}.

However, most existing LLM applications in this domain are limited to single-turn interactions or non-critical tasks. Few studies have successfully demonstrated the capability of LLMs to automate the entire, multi-step process of contingency plan generation, which requires complex reasoning, adherence to strict regulatory guidelines, and the synthesis of structured documents. EpiPlanAgent addresses this gap by focusing on the end-to-end automation of the planning workflow, a critical yet under-explored area in public health AI.

\subsection{Agentic AI Systems for Complex Task Orchestration}
\label{sec: agentic_ai}

Agentic AI systems, which involve autonomous agents making decisions and taking actions without constant human oversight, represent a significant paradigm shift from traditional single-pass LLM applications~\citep{choure2025agentic}. These systems are particularly well-suited for complex, multi-step tasks like emergency response and disaster management, where planning, verification, and execution must be tightly integrated~\citep{leidos}.

Recent proposals, such as PlanAID, have explored combining the benefits of LLMs with symbolic planners to assist in emergency operations planning~\citep{lee2024planning}. Similarly, the concept of a multi-agent system like Disaster Copilot has been proposed to overcome systemic challenges in disaster management by augmenting human-machine intelligence~\citep{li2025disaster}. EpiPlanAgent aligns with this trend by adopting the agentic approach. Specifically, we leverage the \textbf{SigmaFlow} framework, which provides a robust, graph-based orchestration layer for managing task dependencies, parallel execution, and the integration of diverse tools (Model, Code, RAG), ensuring a structured and reliable workflow for plan generation. This contrasts with general-purpose multi-agent frameworks, offering a specialized, flow-centric approach optimized for structured output generation.

\subsection{Retrieval-Augmented Generation (RAG) for Domain Grounding}
\label{sec: rag_related}

A key challenge in deploying LLMs for critical domain applications, such as public health, is ensuring factual accuracy and mitigating hallucinations. RAG has emerged as the leading technique to ground LLMs with authoritative, external knowledge~\citep{wada2025retrieval}. Research in healthcare has demonstrated that RAG can significantly elevate the quality and consistency of LLM outputs by retrieving relevant clinical guidelines and evidence~\citep{ke2025retrieval,ng2025rag}.

For public health, RAG systems have been proposed to enhance question-answering by retrieving data from complementary sources, such as the MEGA-RAG framework~\citep{xu2025mega}. EpiPlanAgent extends the application of RAG by utilizing a highly \textbf{structured, pre-classified knowledge base} of epidemic response actions. Unlike typical RAG systems that rely on vector similarity search over unstructured text, our approach uses the RAG mechanism for \textbf{precise, exact-match retrieval} of structured JSON data based on the identified disease type. This ensures that the retrieved "Candidate Plans" are directly authoritative and immediately usable by the subsequent LLM steps for structured task list generation, maximizing both the recall and precision of the domain knowledge grounding.

\section{Further Discussions and Conclusion}
\label{sec: conclusion}

\subsection{Conclusion}
\label{sec: conclusion_summary}

In this study, we introduced EpiPlanAgent, a novel agent-based system designed for the automated generation of structured epidemic emergency response plans. By integrating the reasoning power of Large Language Models (LLMs) with the robust orchestration capabilities of the SigmaFlow framework and a domain-specific Retrieval-Augmented Generation (RAG) mechanism, EpiPlanAgent addresses the critical need for rapid, consistent, and guideline-aligned contingency planning in public health emergencies.

Our comprehensive evaluation, involving public health professionals and real-world scenarios, demonstrated the system's significant impact:
\begin{itemize}
    \item \textbf{Enhanced Quality:} EpiPlanAgent-assisted plans achieved a significantly higher Completeness Score ($\mathbf{82.4\%}$) compared to manual planning ($\mathbf{68.7\%}$), confirming its ability to ensure the inclusion of essential response actions.
    \item \textbf{Improved Efficiency:} The system dramatically reduced the Plan Generation Time by $\mathbf{61.8\%}$, highlighting its value in time-critical situations.
    \item \textbf{High Alignment:} The strong correlation ($r=0.92$) with expert-authored content validates the system's ability to align with professional judgment and authoritative guidelines.
\end{itemize}
EpiPlanAgent represents a practical and scalable solution that leverages agentic AI to transform traditional, labor-intensive planning workflows into an intelligent, standardized, and efficient digital process, offering a clear blueprint for future AI-driven public health preparedness.

\subsection{Limitations and Future Work}
\label{sec: limitations}

Despite the promising results, EpiPlanAgent currently has several limitations that suggest avenues for future research:

\subsubsection{Limitations}
\begin{itemize}
    \item \textbf{Static Knowledge Base:} The current RAG mechanism relies on a pre-compiled, static knowledge base of 8 diseases. While this ensures high precision, it limits the system's ability to adapt to novel or rapidly evolving pathogens for which structured guidelines do not yet exist.
    \item \textbf{Scenario Complexity:} The current workflow primarily focuses on generating the initial response task list based on a single set of reporting information. It does not yet fully incorporate complex, multi-stage scenario simulation or dynamic re-planning based on real-time data feeds (e.g., changing case numbers, resource depletion).
    \item \textbf{LLM Dependency:} The system's performance is inherently dependent on the underlying LLM's reasoning and instruction-following capabilities, particularly in the critical steps of "Epidemic Type Extraction" and "Task List Generation."
\end{itemize}

\subsubsection{Future Work}
To address these limitations and further advance the field, we propose the following future research directions:
\begin{itemize}
    \item \textbf{Dynamic Knowledge Integration:} We plan to integrate a dynamic RAG component that can perform real-time web searches or query external, frequently updated public health databases to handle emerging threats and ensure the knowledge base remains current.
    \item \textbf{Multi-Agent Simulation and Verification:} Future iterations will explore a more sophisticated multi-agent architecture where specialized agents (e.g., a "Scenario Agent," a "Verification Agent," and a "Resource Allocation Agent") collaborate to simulate the plan's execution and verify its feasibility and impact before finalization.
    \item \textbf{Human-in-the-Loop Refinement:} Developing an interactive interface that allows public health experts to easily review, modify, and provide structured feedback directly into the agentic workflow. This feedback can then be used to fine-tune the LLM prompts and RAG data, creating a continuous learning loop for the system.
    \item \textbf{Cross-Jurisdictional Adaptation:} Expanding the knowledge base to include guidelines from multiple national or international public health organizations to enable the system to generate plans that are adaptable to different jurisdictional requirements.
\end{itemize}
By pursuing these directions, we aim to evolve EpiPlanAgent into a more robust, adaptive, and comprehensive tool for global public health preparedness.

\bibliography{ref}
\bibliographystyle{ref-style}

\clearpage
\onecolumn
\appendix 
\etocdepthtag.toc{mtappendix}
\etocsettagdepth{mtchapter}{none}
\etocsettagdepth{mtappendix}{subsection}
\renewcommand{\contentsname}{Appendix}
\tableofcontents
\clearpage

\section{Domain Knowledge Base Construction}
\label{sec: data}

This appendix provides a detailed description of the structured domain knowledge base used by EpiPlanAgent's Retrieval-Augmented Generation (RAG) mechanism, as discussed in Section \ref{sec: rag}. The knowledge base is crucial for ensuring the authority and standardization of the generated emergency response plans.

\subsection{Knowledge Base Scope and Content}
The knowledge base is stored locally and is organized by disease type. It covers a total of \textbf{8 infectious diseases} across various transmission routes, including:
\begin{itemize}
    \item Pertussis
    \item Hand-Foot-and-Mouth Disease
    \item Dengue Fever
    \item Monkeypox
    \item Cholera
    \item Chikungunya Fever
    \item Influenza
    \item COVID-19
\end{itemize}
Each disease is represented by a dedicated JSON file, which contains a comprehensive list of all potential emergency response actions and their associated metadata.

\subsection{Structured Data Schema}
Unlike traditional RAG systems that rely on unstructured text documents, our knowledge base is highly structured. Each entry in the JSON file represents a single, atomic response action and adheres to a strict schema, which is designed to facilitate precise retrieval and structured output generation by the LLM.

The schema for each response action is defined by the following key-value pairs (translated from the original Chinese fields):

\begin{table}[b!]
\centering
\caption{Schema of the Structured Response Action Knowledge Base}
\label{tab: kb_schema}
\begin{tabular}{|l|p{10cm}|}
\hline
\textbf{Field} & \textbf{Description} \\
\hline
\texttt{Action} & The specific response measure (e.g., "Team Deployment," "EPI," "Lab Testing"). \\
\hline
\texttt{Trigger Condition} & The logical condition (often a boolean expression) that must be met to initiate the action (e.g., "Confirmed OR Suspected Case"). \\
\hline
\texttt{Work Requirement} & Detailed operational guidelines and specifications for executing the action. \\
\hline
\texttt{Responsible Party (A/B Level)} & The institution or department responsible for execution under high-risk (A/B) scenarios. \\
\hline
\texttt{Responsible Party (C/D Level)} & The institution or department responsible for execution under low-risk (C/D) scenarios. \\
\hline
\texttt{Time Limit} & The time constraint within which the action must be completed (e.g., "2 hours," "24 hours," "14 days"). \\
\hline
\texttt{Termination Condition} & The condition that signals the completion or termination of the action. \\
\hline
\end{tabular}
\end{table}

\subsection{Knowledge Base Example}
An excerpt from the Dengue Fever knowledge base demonstrates how the structured data is used to define an action:

\begin{lstlisting}
{
    "Action": "Epidemiological Investigation",
    "Trigger Condition": "Confirmed, Suspected, or Clinically Diagnosed Case",
    "Work Requirements": "Basic information, onset of illness, medical visits, and laboratory testing results. Activity trajectory tracing: Activities during the 14 days prior to onset of illness until mosquito-proof isolation period."
    "Responsible Party A/B": "Municipal CDC Infectious Disease Prevention Institute, District CDC",
    "Responsible Party C/D": "District CDC",
    "Time Limit": "24 hours",
    "Termination Condition": "Submission of Epidemiological Investigation Report"
}
\end{lstlisting}

\subsection{RAG Mechanism and Precision}
The use of this structured format enables the EpiPlanAgent to employ a highly precise RAG mechanism. Instead of relying on vector similarity search over large text blocks, the Agent uses the identified disease type (e.g., "Dengue Fever") as an exact key to retrieve the corresponding JSON file. This approach guarantees:
\begin{itemize}
    \item \textbf{High Recall:} All relevant actions for the specific disease are retrieved.
    \item \textbf{High Precision:} The retrieved data is directly authoritative and free from semantic drift or hallucination, as it is sourced from the pre-verified structured guidelines.
\end{itemize}
The retrieved list of actions, referred to as the "Candidate Plans," is then passed to the subsequent LLM nodes for filtering and synthesis into the final response task list, ensuring the plan is both comprehensive and compliant with public health standards.

\newpage
\section{Prompts Used in EpiPlanAgent}
\label{sec: prompts}

This appendix provides the full set of prompts used by the Large Language Model (LLM) nodes within the EpiPlanAgent's agentic workflow, as orchestrated by the SigmaFlow framework. All original prompts were written in Chinese to ensure optimal performance with the domain-specific knowledge base. The English translations here are provided for clarity.

\subsection{Prompt for Epidemic Type Extraction}
\label{sec: prompt_type_extraction}
This prompt is used in the \textbf{Epidemic Type Extraction} step to identify the specific disease from the candidate list based on the initial reporting information.

\begin{tcolorbox}[
    colback=cyan!5,
    colframe=cyan!40!black,
    title=Prompt 1: Epidemic Type Extraction
]

Based on the epidemic reporting information, select and determine the specific epidemic type from the candidate list, and return the result in JSON format, 
e.g., \{"Epidemic Type": "xxx"\}. \\
Candidate Epidemic Types: \placeholder{candidate\_epidemic\_types} \\
Epidemic Reporting Information: \placeholder{epidemic\_reporting\_information} \\
Answer:

\end{tcolorbox}

\subsection{Prompt for Extracting Condition Points}
\label{sec: prompt_condition_points}
This prompt is used in the \textbf{Extract Condition Points} step to decompose the trigger conditions from the candidate plans into atomic, independent judgment points.

\begin{tcolorbox}[
    colback=cyan!5,
    colframe=cyan!40!black,
    title=Prompt 2: Extracting Condition Points
]

Based on the candidate plans, extract each independent, atomic condition point from the trigger conditions and list them out. Do not repeat or categorize them.
Example: \\
1. Confirmed cases \\
2. Suspected cases \\
3. Clinically diagnosed cases. \\

Candidate Plan - Trigger Conditions: \placeholder{all\_trigger\_conditions} \\
Answer:

\end{tcolorbox}

\subsection{Prompt for Case Structuring}
\label{sec: prompt_case_structuring}
This prompt is used in the \textbf{Case Structuring} step to match the initial reporting information against the atomic condition points, converting unstructured input into structured facts.

\begin{tcolorbox}[
    colback=cyan!5,
    colframe=cyan!40!black,
    title=Prompt 3: Case Structuring
]

Condition Points: \placeholder{condition\_points} \\
Epidemic Reporting Information: \placeholder{epidemic\_reporting\_information} \\

Based on the epidemic reporting information, output the structured analysis according to the condition points. Example: \\
1. Confirmed cases: Yes \\
2. Suspected cases: No \\

Answer:

\end{tcolorbox}

\subsection{Prompt for Task List Generation (Initial)}
\label{sec: prompt_task_initial}
This prompt is used in the \textbf{Task List Generation} step (initial run) to synthesize the final structured response task list.

\begin{tcolorbox}[
    colback=Salmon!5,
    colframe=Salmon!90!Black,
    title=Prompt 4: Task List Generation (Initial)
]

Risk Cases: \placeholder{risk\_cases} \\
Basic Case Information: \placeholder{basic\_case\_information} \\
Structured Epidemic Reporting Information: \placeholder{structured\_info} \\
Candidate Plans: \placeholder{candidate\_plans} \\

Based on all the information provided above, select the next task list from the candidate plans, and strictly output in JSON format: \\
$[$\{"Action": "xxx", "Work Requirement": "xxx", "Responsible Party": "xxx", "Time Limit": "xxx"\}, ...$]$ \\
Answer:

\end{tcolorbox}

\subsection{Prompt for Task List Generation (Iterative)}
\label{sec: prompt_task_iterative}
This prompt is used in the \textbf{Task List Generation} step (iterative run) to refine the task list based on previous execution feedback.

\begin{tcolorbox}[
    colback=Salmon!5,
    colframe=Salmon!90!Black,
    title=Prompt 5: Task List Generation (Iterative)
]

Risk Cases: \placeholder{risk\_cases} \\
Basic Case Information: \placeholder{basic\_case\_information} \\
Structured Epidemic Reporting Information: \placeholder{structured\_info} \\
Candidate Plans: \placeholder{candidate\_plans} \\
Previous Task List and Feedback: \placeholder{previous\_task\_feedback} \\

Based on all the information provided above and the previous task list and feedback, select the next round of the task list from the candidate plans, and strictly output in JSON format:
$[$\{"Action": "xxx", "Work Requirement": "xxx", "Responsible Party": "xxx", "Time Limit": "xxx"\}, ...$]$ \\
Answer:

\end{tcolorbox}

\newpage
\section{Case Study: Dengue Fever Response Planning}
\label{sec: case_studies}

This appendix presents a detailed case study demonstrating the two-round agentic workflow of EpiPlanAgent, as described in Section \ref{sec: workflow}. The case involves a confirmed Dengue Fever case, where the Agent must generate an initial response plan and then iteratively refine it based on the execution feedback. All sensitive information has been de-identified and translated into English for convenience.

\subsection{Case Background and Initial Input}
The case involves Patient Z, a 24-year-old female student residing in District Y, City S, who was diagnosed with Dengue Fever (DENV-2) while visiting J County, Z Province. The case was reported to City S CDC, requiring an immediate investigation as local infection could not be ruled out.

\subsubsection{Initial Input Data}
The Agent receives the following structured input data, which triggers the workflow:
\begin{itemize}
    \item \textbf{Epidemic Reporting Information:} A detailed text report from J County CDC and the initial internal report from City S CDC, confirming the diagnosis and the patient's residence in District Y.
    \item \textbf{Risk Case Information:} Structured metadata indicating the disease is Dengue Fever, the risk level is B, and the required response teams are City-level and District-level.
    \item \textbf{Basic Case Information:} Patient Z's details, including symptom onset (July 26, 2025), current location (X hospital), and initial judgment of "local infection" possibility.
\end{itemize}

\subsection{Round 1: Initial Response Plan Generation}
\label{sec: round1}

\subsubsection{Agentic Workflow Steps}
\begin{enumerate}
    \item \textbf{Epidemic Type Extraction:} The LLM node correctly identifies the disease as \textbf{Dengue Fever} from the candidate list.
    \item \textbf{RAG Retrieval:} The RAG node retrieves the full set of structured Candidate Plans for Dengue Fever from the knowledge base.
    \item \textbf{Case Structuring:} The LLM extracts atomic condition points (e.g., "Confirmed Case," "Local Infection Possibility") and structures the input data against these points (e.g., "Confirmed Case: Yes," "Local Infection Possibility: Yes").
    \item \textbf{Task List Generation:} The LLM synthesizes the structured case facts and the Candidate Plans to generate the initial task list, focusing on immediate, time-critical actions.
\end{enumerate}

\subsubsection{Initial Task List (Excerpt)}
The initial plan contained 13 tasks. A representative excerpt is shown in Table \ref{tab: round1_tasks}. The focus is on rapid deployment, initial investigation, and immediate vector control measures.

\begin{table}[h]
\centering
\caption{Excerpt of Initial Task List (Round 1)}
\label{tab: round1_tasks}
\begin{tabular}{|p{2.5cm}|p{5.5cm}|p{3.5cm}|p{2cm}|}
\hline
\textbf{Action} & \textbf{Work Requirement} & \textbf{Responsible Party} & \textbf{Time Limit} \\
\hline
Team Deployment & District CDC team to be formed and deployed for investigation. & District Y CDC & 2 hours \\
\hline
EPI & Complete preliminary EPI within 24 hours, including activity tracing (14 days prior to isolation). & City CDC, District Y CDC & 24 hours \\
\hline
EPI & Delineate risk areas: Core Area (residence/workplace $\ge$ 100m), Alert Area (Core + 200m). & City CDC, District Y CDC & 24 hours \\
\hline
Vector Sampling & Collect Aedes mosquito samples (adults and larvae) for testing. & District Y CDC & 48 hours \\
\hline
Vector Control & Conduct emergency mosquito control in Core/Alert areas (e.g., ULV spraying, source reduction). & City CDC, District Y CDC & 3 days to show effect \\
\hline
\end{tabular}
\end{table}

\subsection{Round 2: Iterative Refinement Based on Feedback}
\label{sec: round2}

\subsubsection{Feedback from Round 1 Execution}
The Agent receives the initial task list with execution feedback, which becomes the key input for the iterative step. Key feedback points include:
\begin{itemize}
    \item \textbf{EPI Result:} Patient Z had no travel history outside City S in the 14 days prior to onset, increasing the likelihood of local infection.
    \item \textbf{Risk Area Delineation:} Two risk areas were defined: Patient Z's dormitory (University A) and her internship company (Company B).
    \item \textbf{Vector Monitoring Result:} Mosquito monitoring showed high Aedes density in the core area of Company B (8.67 per tent-hour), exceeding the safety threshold.
    \item \textbf{Coordination Result:} Gene sequencing results were received from Z Province.
\end{itemize}

\subsubsection{Iterative Task List Generation}
The Agent enters the iterative branch, using the \textbf{Prompt 5 (Iterative)} which incorporates the `Previous Task List and Feedback`. The LLM's reasoning shifts from initial deployment to advanced investigation and risk management based on the new facts.

\subsubsection{Iterative Task List (Excerpt)}
The refined plan contains 5 tasks, focusing on laboratory confirmation, in-depth EPI, and long-term risk control, as shown in Table \ref{tab: round2_tasks}.

\begin{table}[h]
\centering
\caption{Iterative Task List (Round 2)}
\label{tab: round2_tasks}
\begin{tabular}{|p{2.5cm}|p{5.5cm}|p{3.5cm}|p{2cm}|}
\hline
\textbf{Action} & \textbf{Work Requirement} & \textbf{Responsible Party} & \textbf{Time Limit} \\
\hline
Lab Testing & Compare patient sample gene sequence with local/imported strains. & City CDC Pathogen Institute & 7 days \\
\hline
EPI & Conduct further on-site investigation, focusing on suspected mosquito bite history and areas with high vector density. & City CDC, District Y CDC & 24 hours \\
\hline
Lab Testing & Conduct nucleic acid typing on collected mosquito vector samples. & City CDC Pathogen Institute & 24 hours \\
\hline
Risk Assessment & Conduct further risk assessment based on case investigation, risk control, and gene sequencing results to refine next steps. & City CDC Emergency Office & 24 hours \\
\hline
Risk Personnel Control & Organize community to implement 14-day medical observation for common exposure contacts. & City CDC, District Y CDC & 14 days \\
\hline
\end{tabular}
\end{table}

\subsection{Demonstration of Agentic Value}
This case study clearly demonstrates the value of the EpiPlanAgent's agentic workflow:
\begin{itemize}
    \item \textbf{Contextual Adaptation:} The Agent successfully transitions from generic initial response tasks (Round 1) to highly specific, evidence-based actions (Round 2) by incorporating real-time field feedback (e.g., high mosquito density, local infection possibility).
    \item \textbf{Structured Output:} The output remains strictly structured (JSON format), ensuring immediate usability by downstream systems or human operators.
    \item \textbf{Iterative Refinement:} The workflow's iterative loop allows the Agent to function as a continuous decision-support system, guiding the response team through the evolving stages of the epidemic.
\end{itemize}

\newpage

\end{document}